\documentclass[3p,times]{article}

\usepackage{amssymb}

\usepackage[figuresright]{rotating}

\usepackage[american]{babel}
\usepackage[numbers]{natbib}
\usepackage{mathtools}
\usepackage{booktabs}
\usepackage{tikz}
\usepackage{graphicx, times}
\usepackage{subcaption} 
\usepackage{bm}
\usepackage{url}
\usepackage{amsthm, amssymb, amsfonts, amsmath}
\usepackage{bbold}
\usepackage{dsfont}
\usepackage{ragged2e}
\usepackage{booktabs}
\usepackage{algorithm, algorithmic}
\usepackage{pgfplots}
\usepackage{comment}
\pgfplotsset{compat=1.17}
\usepackage{authblk} 
\usepackage[title]{appendix}
\usepackage[margin=2.5cm]{geometry}

\newcommand{\bd}[1]{\boldsymbol{#1}}
\newcommand{\set}[1]{\mathcal{#1}}
\newcommand{\1}[1]{\mathds{1}(#1)}
\newcommand{\e}[1]{\exp\{#1\}}
\newcommand{\x}{\boldsymbol{x}}

\newcommand{\tht}{\boldsymbol{\theta}}
\newcommand{\Real}{\mathds{R}}

\begin{document}




\title{Risk-based Calibration for Generative Classifiers}

\author[1]{Aritz~P\'erez }
\author[2]{Carlos~Echegoyen}
\author[2]{Guzm\'an~Santaf\'e}

\affil[1]{Machine Learning\\Basque Center for Applied Mathematics\\Bilbao, Spain}
\affil[2]{Spatial Statistics Group\\Public University of Navarre\\Pamplona, Spain}


\date{}
\maketitle

\begin{abstract}
Generative classifiers are constructed on the basis of a joint probability distribution and are typically learned using closed-form procedures that rely on data statistics and maximize scores related to data fitting. However, these scores are not directly linked to supervised classification metrics such as the error, i.e., the expected 0-1 loss. To address this limitation, we propose a learning procedure called risk-based calibration (RC) that iteratively refines the generative classifier by adjusting its joint probability distribution according to the 0-1 loss in training samples. This is achieved by reinforcing data statistics associated with the true classes while weakening those of incorrect classes. As a result, the classifier progressively assigns higher probability to the correct labels, improving its training error. Results on 20 heterogeneous datasets using both naïve Bayes and quadratic discriminant analysis show that RC significantly outperforms closed-form learning procedures in terms of both training error and generalization error. In this way, RC bridges the gap between traditional generative approaches and learning procedures guided by performance measures, ensuring a closer alignment with supervised classification objectives.
\end{abstract}




\section{Introduction} \label{sec:intro}
Supervised classification is one of the most crucial problems in machine learning, involving the construction of a classifier from training data that minimizes the error, i.e., the expected 0-1 loss, where the 0-1 loss assigns 1 to misclassified instances and 0 to correctly classified ones. A classifier essentially is a function that maps instances to a set of class labels. Typically, classifiers are expressed in a parametric form, and their parameters are obtained using a learning algorithm from training data consisting of a set of labeled instances drawn i.i.d. from an unknown true probability distribution.

Depending on the modeling approach, classifiers can be categorized into generative and discriminative~\cite{Jebara12, Zheng23}. 
Discriminative classifiers learn a direct mapping from instances to class labels by optimizing a loss function closely related to the error. Typically, an appropriate parametric family of classifiers and a suitable loss function that upper bounds the 0-1 loss are chosen. Then, following the empirical risk minimization principle, the parameters are estimated by minimizing the average loss on the training data as the empirical surrogate of the true expected loss~\cite{hastie2009, vapnik2013nature}. For example, support vector machines optimize the average hinge loss, thereby directly targeting the minimization of classification-related scores.

Alternatively, generative classifiers address supervised classification by first approximating the unknown joint probability distribution over instances and class labels~\cite{Jebara12}. Then, they construct the mapping from instances to labels by selecting, for each instance, the label that minimizes the 0-1 loss w.r.t. the modeled joint probability distribution. Unlike discriminative methods, which optimize classification-related scores such as the empirical surrogates of the error, generative classifiers focus on probabilistic modeling of the domain by maximizing measures of data fitness, such as the maximum likelihood or maximum a posteriori criteria. This category includes models from the exponential family, such as Quadratic Discriminant Analysis (QDA), as well as classifiers based on Bayesian networks under various assumptions, including discrete Bayesian networks~\cite{bielza2014discrete} and conditional Gaussian networks~\cite{perez2006gaussian}. 

Unfortunately, generative classifiers often perform worse from a classification perspective than discriminative classifiers~\cite{ng2001discriminative}. Learning by maximizing likelihood-related scores instead of directly optimizing a classification-related score often results in classifiers with higher error. For instance, it is known that the direct impact of the likelihood function on classification performance can become negligible as the dimensionality increases~\cite{friedman97}.

However, generative classifiers remain an attractive approach to supervised classification due to several key practical advantages over discriminative classifiers, as they construct a model of the joint probability distribution of the domain. Typically, the joint probability distribution belongs to a parametric family, where the parameters provide intuitive insights into the domain, such as the mean vector and covariance matrix. Furthermore, generative classifiers facilitate the incorporation of prior knowledge and domain expertise through prior distributions in a Bayesian framework, particularly when a conjugate prior is known. They offer comprehensive data modeling by capturing the joint probability distribution of input features and class labels, leading to a better understanding of the data-generating process \cite{augenstein2019generative,GM20}.

Generative classifiers also benefit from their inherently constrained capacity when the number of parameters is low, reducing the risk of overfitting and ensuring stable parameter updates. Their limited model complexity prevents them from fitting noise in the training data, allowing adjustments to remain aligned with the true data distribution. Moreover, their probabilistic structure imposes a natural form of regularization, further enhancing their generalization capabilities while maintaining robustness in real-world applications. Moreover, they often achieve strong performance with smaller training set sizes than discriminative approaches, as they reach their best performance more quickly, possibly with training sizes logarithmic in the number of parameters \cite{ng2001discriminative}.

Additionally, they can generate synthetic data samples by sampling from the joint probability distribution \cite{Cuesta2019}, which aids in data augmentation and anomaly detection. Generative classifiers also handle missing data effectively using the expectation-maximization algorithm \cite{dempster1977}, enhancing robustness in practical scenarios \cite{Tolou23,Kim23}. Their compatibility with Bayesian decision theory allows them to optimize decisions under uncertainty \cite{Murphy12}. Finally, when the class-conditional distribution is accurately modeled within the joint distribution, generative classifiers can provide optimal predictions for a given cost-sensitive loss function without requiring further adjustments \cite{elkan2001costsensitive}.

The limitations in classification performance, coupled with the advantages of modeling a full probability distribution, have led to methods that guide generative classifier learning using classification-related scores, such as minimizing the average log-loss in training. In this line, ~\cite{greiner02, roos2005discriminative, pernkopf2005discriminative, su2008discriminative} developed methods to learn discrete Bayesian networks from a discriminative perspective by using gradient descent (GD) to minimize the conditional log likelihood, that is proportional to average log-loss in training. 

Unfortunately, GD could suffer several drawbacks, particularly those concerning constrained parameters. Violations of these constraints may yield invalid parameter values, numerical instability, and reduce model performance. Techniques like projection, parameter transformation~\cite{greiner02, roos2005discriminative}, or Lagrange multipliers~\cite{bertsekas1997nonlinear} are indispensable for enforcing constraints but may introduce convergence challenges. Besides, these techniques may entail increasing the computational cost of each iteration of GD; for instance, transformations applied to the covariance matrix to fulfill constraint requirements often require computing its inverse, a computationally intensive operation that can significantly limit practical applications in high-dimensional domains. Therefore, optimizing generative classifiers under constrained parameters requires a gradient descent approach that is specifically tailored to both the probability model and the parameter constraint-handling mechanisms. This adaptation can complicate implementation and potentially limit efficiency and effectiveness, making specialized methods imperative to mitigate these challenges and ensure reliable learning of generative classifiers.

In this work, we present risk-based calibration (RC), a simple, efficient, and effective iterative method for learning generative classifiers with strong classification performance. RC calibrates the probability model of the generative classifiers guided by the training error by updating data statistics. It begins with initial parameter estimates computed using the training data statistics, and then iteratively refines these statistics using a soft version of the 0-1 loss in training samples. This produces an updated probability model tailored to deal with the classification problem. The data statistics are adjusted using a probabilistically labeled version of the training data, and the parameters are derived through an existing data-statistics-based learning algorithm, ensuring their validity without requiring additional transformations to enforce parameter constraints. We empirically show classification improvement obtained with RC through extensive experimental results using two generative classifiers on 20 heterogeneous datasets, with the number of instances ranging from 150 to 70000 and feature dimensions between 4 and 512.

The remainder of this paper is organized as follows. Section~\ref{sec:background} provides a brief introduction to supervised classification. Section~\ref{sec:generative} presents generative classifiers and closed-form learning algorithms, introducing naïve Bayes and quadratic discriminant analysis as representative examples. Section~\ref{sec:proposal} proposes Risk-Based Calibration (RC) for learning generative classifiers, guided by a soft version of the training error. Section~\ref{sec:related} discusses the connections between RC and the two most closely related methods for learning generative classifiers. Section~\ref{sec:experiments} empirically evaluates RC in terms of training and generalization error. Finally, Section~\ref{sec:conclusion} summarizes the main contributions of this work. Additionally, the appendix provides details on the implementation of the gradient descent method for learning generative classifiers from the exponential family, the maximum a posteriori parameter mapping for naïve Bayes and quadratic discriminant analysis, and further experiments analyzing RC.

\section{Supervised classification}
\label{sec:background}

Let $\set{X} \subset \Real^d$ and $\set{Y}=\{1, \cdots, r\}$ be the space of instances and the set of class labels, respectively. A classifier $h$ is a function from instances to labels, $h: \set{X} \rightarrow \set{Y}$, and the set of classifiers is denoted by $\set{H}$. The loss function measures the discrepancy between the predicted class labels and the true class labels. The loss of a classifier $h$ evaluated at instance $(\x,y)$ is a function $l: \set{H} \times (\set{X},\set{Y}) \rightarrow [0,\infty)$. The natural loss in classification is the 0-1 loss, $l_{01}(h,(\x,y))=\1{h(x)\neq y}$, where $\1{\cdot}$ is an indicator function that outputs one for a true argument and zero otherwise. The goal of supervised learning can be defined as selecting the classifier $h \in \set{H}$ that minimizes the expected $0-1$ loss (\emph{error}):
$$\min_{h \in \set{H}} E_{p^*} [l_{01}(h,(\x,y))],$$ 
where $p^* \in \Delta(\set{X},\set{Y})$ is the underlying (unknown) distribution of the data. The classifier that minimizes the error is named the \textit{Bayes classifier}.

In standard supervised classification settings, the underlying data distribution $p^*$ is unknown, and we only have access to a training set \((X,Y) \in \set{X}^m \times \set{Y}^m\), where \((X,Y)=\{(\mathbf{x}^i, y^i)\}_{i=1}^m\) are i.i.d. samples drawn from \(p^*\). In this scenario, the expected loss is usually approximated by the average loss on the training data. An appropriate loss function is then chosen so that, is an upper bound of the 0-1 loss and in combination with the parametric form of the selected classifier family, the minimization of its average in training becomes tractable. This approach enables the development of efficient learning algorithms that leverage both the parametric form of the chosen classifier family and the surrogate to the error. 

\section{Generative classifiers}\label{sec:generative}

Generative classifiers are constructed upon a joint probability distribution of the instances and class labels, $p_h \in \Delta(\set{X}\times \set{Y})$,
\begin{equation}
h(\x) := \arg \max_y p_h(\x,y) = \arg \max_y p_h(y|\x) 
\end{equation}
Generative classifiers are fundamentally motivated by their ability to represent the Bayes classifier, assuming they accurately capture the true class' conditional distribution, $p_h(y|\x)= p^*(y|\x)$ for $\x \in \set{X}$.  
They focus the learning on obtaining a good estimate of the joint distribution $p^*(\x,y)$ from training data, and thus, they optimize scoring functions indirectly related to classification. Typically, generative classifiers are learned by finding the parameters that maximize the likelihood of the training data. In contrast, this work proposes a learning procedure for generative classifiers guided by the training error, i.e., the average 0-1 loss in training data. 

One of the main advantages of generative classifiers under certain parametric assumptions is that they have closed-form algorithms for obtaining the maximum likelihood or maximum a posteriori parameters. We say that a learning algorithm, $a$, has a closed-form when it is defined as the composition of a (sufficient) statistics mapping function
$$
s: \set{X}^m \times \set{Y}^m \rightarrow \Real^k,$$
and a parameter mapping function
$$
\theta: \Real^k \rightarrow \Theta,
$$
that is,
$$
a := \theta \circ s.
$$
The statistics mapping summarizes all the relevant information from the training data $(X, Y)$ into $k$ statistics, $\bd{s}$, which serve as sufficient statistics for the model parameters, so the training data is independent of the parameters of the model given the statistics. These statistics capture all the necessary information in a compact form and are then used to compute the parameters of the generative classifier analytically. Generative classifiers based on the exponential family, for example, have closed-form algorithms that maximize the likelihood of the training data, with the statistics mapping being closely related to moment statistics.

From here on, we consider that the statistics mapping involved in the closed-form learning algorithm is additively decomposable, i.e., for $(X,Y) \in \set{X}^m,\times \set{Y}^m$, we have that $\bd{s} =s(X,Y)=\sum_{\x,y \in X,Y} s(\x,y)$, where with a slight abuse in the notation $s(X,Y)$ and $s(\x,y)$ represent the statistics mapping over a training set $(X,Y)$ and over a labeled instance $(\x,y)$, respectively.

\subsection{Examples of generative classifiers and learning algorithms}\label{sec:examples_generative}

In this work, we consider Naïve Bayes (NB) for categorical variables and quadratic discriminant analysis (QDA) for continuous variables as the representative generative classifiers. The statistics mapping for both NB and QDA can be given by the class one-hot encoding $s(\x,y)= (\1{y=1}\cdot \psi(\x),\cdots ,\1{y=r}\cdot \psi(\x))$, where $\psi(\x)$ extracts the statistics from the features that typically correspond to those required to compute the zeroth, first, and second moments.

In the case of NB with discrete variables, the $i$-th input feature, $x_i$, has support $\set{X}_i=\{1,...,r_i\}$ for $i=1,...,n$. NB assumes that the input features are independent given the class variable, $p_h(\x,y)= p_h(y)\prod_{i=1}^n p_h(x_i|y)$, which leads to the classification rule for $\x$:
$$h(\x)= \arg \max_y p_h(y)\cdot \prod_{i=1}^n p_h(x_i|y),$$
where $p_h(y)$ is the probability distribution of the class label $y$ and $p_h(x_j|y)$ is the probability of $i$-th input variable taken the value $x_i$ given the class label $y$ for $i=1,\cdots,n$. These marginal and conditional distributions are assumed to be categorical, and their parameters can be given by the maximum likelihood estimates obtained from the counting statistics. In particular for NB $\psi(\x)=(1,\1{1=x_1},...,\1{r_1=x_1},...,\1{1=x_n},...,\1{r_n=x_n})$. For NB the statistics obtained from data are $\bd{s}= s(X,Y)=(s_{0,1},s_{1,1,1},...,s_{1,r_1,1},...,s_{n,r_n,1},...,s_{0,r},s_{1,1,r},...,s_{1,r_1,r},...,s_{n,r_n,r})$, where $s_{0,y'}=\sum_{y \in Y}\1{y'=y}$ is the counting statistics associated to the class label $y' \in \set{Y}$ and $s_{i,x'_i,y'}\sum_{x,y \in X,Y}\1{y'=y}\cdot\1{x'_i=x_i}$ is the counting statistics associated to the value $x'_i \in \set{X}_i$ of input feature $i$ given the label $y \in \set{Y}$ for $i=1,...,d$. Then the maximum likelihood parameter mapping $\theta(\bd{s})$ is given by $p_h(y)=s_{0,y}/\sum_{y' \in \set{Y}}s_{0,y'}$ and $p_h(x_i|y)=s_{i,x_i,y}/\sum_{x_i' \in \set{X}_i}s_{i,x_i',y}$ for $y \in \set{Y}$,  $x_i=1,...,r_i$ and $i=1,...n$.

The QDA's classification rule is given by: 
$$h(\x)= \arg \max_y p_h(y)\cdot |\Sigma_y|^{-1/2}\cdot exp\{(\x-\bd{\mu}_y)^T \Sigma_y^{-1} (\x-\bd{\mu}_y)\},$$
where $\bd{\mu}_y$ and $\Sigma_y$ are the mean vector and covariance matrix of $\x$ given the class $y$. This classifier is learned by estimating the maximum likelihood mean vector and covariance matrix given the class using the statistics mapping given by $\psi(\x)=[1,\x, \x^2]$, where $[\cdots]$ rearranges its arguments to conform to a column vector, and $\x^2=\x \cdot \x^T$ is a $n \times n$ matrix. The statistics obtained from data are $\bd{s}=s(X,Y)= [s_{0,1},\bd{s}_{1},\bd{s}^2_1,...,s_{0,r},\bd{s}_{r},\bd{s}^2_r]$ with $\bd{s}_{y'}=\sum_{\x,y \in X,Y}\1{y'=y}\cdot \bd{x}$, and $\bd{s}^2_{y'}=\sum_{\x,y \in X,Y}\1{y'=y}\cdot \x^2$ for $y' \in \set{Y}$. The statistics associated to $1$, $\x$, and $\x^2$ given $y$ are used to get the maximum likelihood estimation for $p_h(y)$, $\mu_y$, and $\Sigma_y$ using the parameter mapping $\theta(\bd{s})$ which consists of $p_h(y)=s_{0,y}/\sum_{y' \in \set{Y}}s_{0,y'}$, $\bd{\mu}_y= \bd{s}_y/s_{0,y}$ and $\Sigma_y= \bd{s}^2_y/s_{0,y} - \bd{\mu}_y \cdot \bd{\mu}_y^T$, for each class label $y \in \set{Y}$.

Using the statistics mapping described here, we can also adapt the parameter mapping for NB and QDA to create a closed-form algorithm that maximizes the a posteriori distribution given the corresponding conjugate priors (see \ref{app:MAP} for further details).

\section{Risk-based calibration} \label{sec:proposal}

The risk-based calibration algorithm (RC) is an iterative heuristic method to improve the error of a generative classifier using a closed-form learning algorithm. This method relies on the very basic intuition of modifying the statistics used by the closed-form learning algorithm, guided by a soft version of the 0-1 loss for each instance. We define the soft 0-1 loss of the generative classifier $h$ in $(\x,y)$ as follows: 
\begin{equation}
l_{s01} (h,(\x,y)) = 1-p_h(y|\x)= \sum_{y'\neq y} p_h(y'|\x).\label{eq:l_s01}
\end{equation}
This loss corresponds to the expected loss of a randomized classifier that selects label \( y \) with probability \( p_h(y|\x) \) for \( y \in \set{Y} \). It enables a smoother calibration of the joint distribution that defines the generative classifier, compared to using the 0-1 loss.

We aim at finding the parameters $\tht^*$ that minimize the average soft 0-1 loss in training (training soft error). Given a training set $(X,Y)$, the training soft error is zero when $p_h(y|\x)=1$ for all $\x,y \in X,Y$. Intuitively, it is possible to lead the statistic $\bd{s}$, and thus the model parameters $\theta(\bd{s})$, towards the optimal classifier by strengthening the statistics $s(\x,y)$ and weakening $s(\x,y')$ for $\x,y \in X,Y$ and every $y' \in \set{Y}$ with $y' \neq y$. The strengthening-weakening update is given by the classifier's soft 0-1 loss at point $(\x,y)$ (see Eq.~\ref{eq:l_s01}). Given the classifier $h$, we propose the following updating rule of the statistics $\bd{s}$ given $(\x,y)$:
\begin{equation}
\bd{s}= \bd{s} + s(\x,y) - \sum_{y' \in \set{Y}} p_h(y'|\x) \cdot s(\x,y').
\end{equation}
The calibration of the statistics $\bd{s}$ is performed by adding $s(\x,y)$ with a weight $1 - p_h(y|\x)$, and by subtracting $s(\x,y')$ with a weight $p_h(y'|\x)$ for all $y'\neq y$. Intuitively, the calibration with $(\x,y)$ increases $p_h(y|\x)$ while decreasing $p_h(y'|\x)$ for $y' \neq y$. Given the data $(X,Y)$ and the classifier $h$, and due to the additive nature of the statistics mapping, $s$, the updating rule of the statistics is simply given by:
\begin{equation}
\bd{s}= \bd{s} + s(X,Y) - s(X,h),\label{eq:rule} 
\end{equation} 
where $s(X,h)= \sum_{\x \in X} \sum_{y' \in \set{Y}} p_h(y'|\x)\cdot s(\x,y')$ is the probabilistic estimate of the statistics given the joint probability distribution of the classifier, $p_h$.

\begin{algorithm}[t!]
\caption{Risk-based Calibration}
\label{alg:RC}
\begin{algorithmic}[1]
\STATE \textbf{Input:} Training set $(X,Y)$ with $m$ labeled instances; learning algorithm $a = \theta \circ s$, defined by the statistics mapping $s$ and the parameter mapping $\theta$; learning rate $lr$.
\STATE \textbf{Output:} Classifier $h$.
\STATE $\bd{s}^{(0)} \gets s(X,Y)$
\STATE $h^{(0)} \gets \theta(\bd{s}^{(0)})$
\STATE $\epsilon^{(0)} \gets \frac{1}{m}\sum_{(\x,y)\in (X,Y)} 1 - p_{h}^{(0)}(y|\x)$
\STATE $t \gets 1$
\REPEAT
    \STATE $\bd{s}^{(t)} \gets \bd{s}^{(t-1)} + lr \cdot \big(s(X,Y) - s(X, h^{(t-1)})\big)$
    \STATE $h^{(t)} \gets \theta(\bd{s}^{(t)})$
    \STATE $\epsilon^{(t)} \gets \frac{1}{m}\sum_{(\x,y)\in (X,Y)} 1 - p_{h}^{(t)}(y|\x)$
    \STATE $t \gets t+1$
\UNTIL{$\epsilon^{(t)} > \epsilon^{(t-1)}$}
\RETURN $h^{(t-1)}$
\end{algorithmic}
\end{algorithm}

The RC procedure is described in Algorithm \ref{alg:RC}. Essentially, RC consists of the iterative application of the calibration of the statistics given in Eq.~\eqref{eq:rule} (step 8) and the update of the model's parameters based on these statistics (step 9).
The algorithm terminates when the training soft error deteriorates (step 12). 
The initialization of the statistics (step 1) can be arbitrary, as long as the statistics remain consistent and produce valid parameters. However, we recommend initializing them using $s(X,Y)$. The statistics obtained from training data provide a more competitive starting point from the empirical error point of view. However, it is possible to try different runs using bootstrap samples from the training data to avoid the convergence to a poor local optimum.

A crucial property of RC is that the sample size of the updated statistics remains invariant to iterations. This is because the strengthening-weakening strategy comes from the soft 0-1 loss (Eq.~\ref{eq:l_s01}) and satisfies $\sum_{y' \in \set{Y}}\1{y=y'}-p_h(y'|\x)= 0$ for any $\x,y \in \set{X},\set{Y}$. This is in contrast to alternative iterative procedures such as Discriminative Frequency Estimate (see Section \ref{sec:dfe}).

\begin{figure}[t!]
    \centering
    \begin{subfigure}{0.49\textwidth}\centering
\begin{tikzpicture}
  \begin{axis}[
    domain=-1:4.5,
    samples=100,
    xlabel={$x$},
    ylabel={},
    xmin=-1, xmax=4.5,
    ymin=0, ymax=1,
    width=1\textwidth,
    height=0.7\textwidth,
    legend style={at={(0.97,0.80)}, anchor=north east},
    grid=both,
    clip=false,
  ]
    \draw[fill=blue, fill opacity=0.2, draw=none] (axis cs:-1,0) rectangle (axis cs:1.25,1);
    \draw[fill=red, fill opacity=0.2, draw=none] (axis cs:1.25,0) rectangle (axis cs:4.5,1);
    
    \addplot[blue, thick] {0.5*(1/sqrt(2*3.14159))*exp(-0.5*(x-0)^2)};
    \addlegendentry{$p(x,y=0)$};
    
    \addplot[red, thick] {0.5*(1/sqrt(2*3.14159))*exp(-0.5*(x-2.5)^2)};
    \addlegendentry{$p(x,y=1)$};
    
    \addplot[blue, dashed, thick] 
      {(0.5*(1/sqrt(2*3.14159))*exp(-0.5*(x-0)^2)) / ((0.5*(1/sqrt(2*3.14159))*exp(-0.5*(x-0)^2)) + (0.5*(1/sqrt(2*3.14159))*exp(-0.5*(x-2.5)^2)))};   
    \addlegendentry{$p(y=0|x)$};      
    
    \addplot[red, dashed, thick] 
      {(0.5*(1/sqrt(2*3.14159))*exp(-0.5*(x-2.5)^2)) / ((0.5*(1/sqrt(2*3.14159))*exp(-0.5*(x-0)^2)) + (0.5*(1/sqrt(2*3.14159))*exp(-0.5*(x-2.5)^2)))};  
    \addlegendentry{$p(y=1|x)$};
    
    \addplot[black, very thick] coordinates {(1.25,0) (1.25,1)};
    
    \draw[blue, dashed] (axis cs:0,0) -- (axis cs:0,1);
    \node[blue, anchor=south] at (axis cs:0,1) {$\mu_1$};
    
    \draw[red, dashed] (axis cs:2.5,0) -- (axis cs:2.5,1);
    \node[red, anchor=south] at (axis cs:2.5,1) {$\mu_2$};
    
    \addplot[only marks, mark=*, mark size=2.5, blue] coordinates {(0,0)};
    \addplot[only marks, mark=*, mark size=2.5, red] coordinates {(1,0)};
    \addplot[only marks, mark=*, mark size=2.5, red] coordinates {(4,0)};
  \end{axis}
\end{tikzpicture}
\caption{The Gaussian mixture model with the initial parameters $\mu_1= 0.0$ and $\mu_2=2.5 $ obtained from the statistics $\bd{s}= (1.0, 0.0, 2.0, 5.0)$. The model fails to classify $x=1$, where the class conditional distribution is $h(\cdot|x=3)=(0.65, 0.35)$ and a soft 0-1 loss is $0.35$.}\label{fig:example.initialization}
\end{subfigure}
    \hfill
\begin{subfigure}{0.49\textwidth}
        \centering
\centering
\begin{tikzpicture}
  \begin{axis}[
    domain=-1:4.5,
    samples=100,
    xlabel={$x$},
    ylabel={},
    xmin=-1, xmax=4.5,
    ymin=0, ymax=1,
    width=1\textwidth,
    height=0.7\textwidth,
    legend style={at={(0.97,0.80)}, anchor=north east},
    grid=both,
    clip=false,
  ]
    \draw[fill=blue, fill opacity=0.2, draw=none] (axis cs:-1,0) rectangle (axis cs:0.92,1);
    \draw[fill=red, fill opacity=0.2, draw=none] (axis cs:0.92,0) rectangle (axis cs:4.5,1);
    
    \addplot[blue, thick] {0.5*(1/sqrt(2*3.14159))*exp(-0.5*(x-(-0.472))^2)};
    
    \addplot[red, thick] {0.5*(1/sqrt(2*3.14159))*exp(-0.5*(x-2.311)^2)};
    
    \addplot[blue, dashed, thick] 
      {(0.5*(1/sqrt(2*3.14159))*exp(-0.5*(x-(-0.472))^2)) / ((0.5*(1/sqrt(2*3.14159))*exp(-0.5*(x-(-0.472))^2)) + (0.5*(1/sqrt(2*3.14159))*exp(-0.5*(x-2.311)^2)))}; 
    
    \addplot[red, dashed, thick] 
      {(0.5*(1/sqrt(2*3.14159))*exp(-0.5*(x-2.311)^2)) / ((0.5*(1/sqrt(2*3.14159))*exp(-0.5*(x-(-0.472))^2)) + (0.5*(1/sqrt(2*3.14159))*exp(-0.5*(x-2.311)^2)))}; 
    
    \addplot[black, very thick] coordinates {(0.92,0) (0.92,1)};
    
    \draw[blue, dashed] (axis cs:-0.472,0) -- (axis cs:-0.472,1);
    \node[blue, anchor=south] at (axis cs:-0.472,1) {$\mu_1$};
    
    \draw[red, dashed] (axis cs:2.311,0) -- (axis cs:2.311,1);
    \node[red, anchor=south] at (axis cs:2.311,1) {$\mu_2$};
    
    \addplot[only marks, mark=*, mark size=2.5, blue] coordinates {(0,0)};
    \addplot[only marks, mark=*, mark size=2.5, red] coordinates {(1,0)};
    \addplot[only marks, mark=*, mark size=2.5, red] coordinates {(4,0)};
  \end{axis}
\end{tikzpicture}
\caption{The model after the first iteration with parameters $\mu_1= -0.47$ and $\mu_2= 2.31$ obtained from the updated statistics $\bd{s}= (0.69, -0.33,  2.31,  5.33)$. The model correctly classifies $x=1$ with label $2$, where the class conditional distribution is $h(\cdot|x=1)=(0.44, 0.56)$ and a soft 0-1 loss is $0.44$. }\label{fig:example.iteration}
\end{subfigure}
    \caption{Example of the initialization and first iteration of RC using a simple model with two parameters $\mu_1$ and $\mu_2$ for unidimensional instances with three training samples} 
    \label{fig:Example}
\end{figure}

\subsection{Example using a simple generative model} 
This is a toy example to provide insights into the behavior of RC using a very simple model and very simple training set with three univariate instances and two class labels.
The training data consist of the three univariate instances $X = (0, 1, 4)$, with the corresponding class labels $Y = (1, 2, 2)$. The generative model is a simplified version of a mixture of Gaussian distributions with class-dependent means, $\mu_1$ and $\mu_2$, and constant weights ($p(y=1)=p(y=2)=0.5$) and variances ($\sigma_1^2 = \sigma_2^2 = 1)$. The joint probability distribution of the generative classifier is defined as
$$p_h(x,y) =\frac{0.5}{\sqrt{2\pi}}\exp\{-(x-\mu_y)^2\}.$$
This simplified model, with only two free parameters, is unable to capture the differences in label proportions or the spread of data across different labels. The learning algorithm is defined by the following statistics mapping
$$s(x,y) = \Bigl[\1{y=1} \cdot (1,x), \1{y=2} \cdot (1,x)\Bigr],$$
and the parameter mapping $\theta(\mathbf{s}) = (\mu_1 = s_2/s_1, \mu_2 = s_4/s_3)$. For the toy experiment we consider a learning rate of $0.5$ for the RC. 

Figure \ref{fig:example.initialization} shows the initial model with parameters $\mu_1=0.0$ and $\mu_2=2.5$ obtained from the statistics $\bd{s}=s(X,Y)=(1.00, 0.00, 2.00, 5.00)$. The decision boundary of the model is at $x=1.25$ and missclasify $x=1$ into the class $1$ (blue region). The soft $0-1$ loss at $x=1$ is $0.65$, and RC take advantage of this information to update the statistics. In Figure \ref{fig:example.iteration} we show the model after the first iteration of RC. The model has parameters $\mu_1=-0.47$ and $\mu_2=2.31$ obtained from the updated statistics $\bd{s}=(0.69, -0.33,  2.31,  5.33)$, where $s(X,Y)-s(X,h)=(1.0, 0.0, 2.0, 5.0) - (1.61, 0.66, 1.39, 4.34)$ summarizes the information given by the training soft error. In particular, the statistics for the previously missclassified point $(x,y)=(1,2)$ has changed from $s(1,2)= (0.00,0.00,1.00,1.00)$ to $s(1,2)+s(1,2)-s(1,h(\cdot|1))= (-0.33, -0.33, 1.33, 1.33)$. The new model correctly classifies all the instances, however it still has a training soft error of $0.17$. RC will take advantage of this information to still produce a better calibrated generative classifier, as it is shown in Table~\ref{table:example.evolution}.

\begin{table}[ht]
\centering
\begin{tabular}{c|cccccccc}
\hline
$x,y$ & $h^{(0)}$ & $h^{(1)}$ & $h^{(2)}$ & $h^{(4)}$ & $h^{(8)}$ & $h^{(16)}$ & $h^{(32)}$ & $h^{(64)}$\\
\hline
0,1 & 0.95 & 0.93 & 0.87& 0.84& 0.86 & 0.90 & 0.93 & 0.96\\
1,2  & 0.35 & 0.56 & 0.80& 0.89 &0.90 & 0.92 & 0.94 & 0.96\\
4,2 & 1.00 & 1.00 & 1.00& 1.00 &1.00 & 1.00 & 1.00 & 1.00\\
\hline
soft &  0.23 & 0.17& 0.11& 0.09& 0.08 & 0.06 & 0.04 & 0.03
\end{tabular}
\caption{The evolution of $h^{(t)}(y|x)$ and its training soft error for different iterations of the RC, $t$, where $t=0$ represents the initial model.}\label{table:example.evolution}
\end{table}

\subsection{Computational complexity of RC} 

The computational cost of each iteration is given by the statistics mapping $s(\cdot)$ and parameter mapping $\theta(\cdot)$. The computational complexity of the statistics mapping is linear in the number of training samples $m$, the dimension of each instance $n$, the dimension of the statistics $k$ and the number of operations required to compute $p_h(y|x)$ for each training sample $o$, $\set{O}(m \cdot (n+ k+ o))$, while the computation of the parameter mapping is independent of $m$ and is usually linear in the number of statistics and parameters $q$, $\set{O}(k + q)$. Due to the additively decomposable assumption for the statistics mapping, it is possible to speed up RC by using stochastic and minibatch approaches, however this is out of scope of this work.

\section{Connections between RC and other methods} \label{sec:related}

The RC is a general-purpose algorithm for learning probabilistic classifiers that presents some connections to other existing methods proposed in the literature. This section shows the relation of RC with two iterative learning procedures: Discriminative Frequency Estimate for learning classifiers based on discrete Bayesian networks~\cite{su2008discriminative}, and the TM algorithm for generative classifiers from the exponential family~\cite{TM_Lauritzen,Guzman07}.

\subsection{Discriminative Frequency Estimate}\label{sec:dfe}

Discriminative Frequency Estimate (DFE)~\cite{su2008discriminative} is used for learning the parameters of classifiers based on discrete Bayesian networks~\cite{bielza2014discrete}, such as na\"ive Bayes (NB). The main motivation for DFE is that learning discrete Bayesian networks based on gradient descent over the average log loss in training~\cite{greiner02} is computationally demanding. DFE is an iterative procedure where, at each iteration, the statistics used to derive the parameters of a discrete Bayesian network classifier are updated. Using our notation, the updating rule is given by:
\begin{align}
\bd{s}= \bd{s} + (1-p_h(y|\x))\cdot s(\x,y)\nonumber  
\end{align}
This corresponds to a heuristic that strengthens the statistics $s(\x,y)$ according to the soft error in training. The main difference from RC in the context of discrete Bayesian network classifiers is that DFE does not weaken the incorrect statistics associated with the instance $(\x,y)$, that is $s(\x,y')$ for $y' \neq y$. An important negative consequence is that DFE increases the equivalent sample size of the statistics 
$\bd{s}$, at each iteration by the training soft error of the classifier obtained in the previous iteration. This can have dramatic effects with large training sets or when the number of iterations for convergence is large. The main limitation of DFE is that it only applies to discrete Bayesian network classifiers.

In~\ref{app:DFE}, we present an experimental comparison between RC and DFE for learning NB as an example of the discrete Bayesian network family of classifiers. As shown in Table~\ref{table:comparison_dfe}, RC achieves lower training and test errors than DFE in 13 out of 20 datasets, performs equally in 2 datasets, and underperforms in 5 datasets. Notably, although DFE was specifically developed for parameter learning in discrete Bayesian network classifiers, RC demonstrates superior performance in most cases, highlighting its effectiveness for learning this family of generative classifiers.

\subsection{The connection between RC and TM}\label{sec:TM}
The TM algorithm~\cite{TM_Lauritzen, sundberg2002convergence} is a general method designed to maximize the conditional likelihood using the parameters of a joint distribution over features and class labels from the exponential family. Conditional likelihood is proportional to the average log loss in training. TM follows an iterative approach based on the maximum likelihood learning algorithm. It is a general algorithm that can be used for regression (\(\set{Y} \subset \mathbb{R}\), continuous) and classification (\(\set{Y} = \{1, \dots, r\}\), categorical).

At each iteration $t$, the TM solves two steps:
\begin{itemize}
    \item T-step: Compute the gradient of the marginal log-likelihood with respect to the parameters: $$\frac{\delta LL(X;\tht)}{\delta \tht},$$
    where $LL(X,\tht)= \sum_{\x \in X} \log \sum_{y \in \set{Y}} p_h(\x,y;\tht)$ is the marginal log likelihood.
    \item M-step: solve the maximization problem.
    $$\tht^{(t+1)}= \arg \max_{\tht} LL(X,Y;\tht) - \frac{\delta LL(X;\tht^{(t)})}{\delta \tht}\cdot \tht,$$where $LL(X,Y,\tht)= \sum_{(\x,y) \in (X,Y)} \log p_h(\x,y;\tht)$ is the log likelihood of the training set and $\delta LL(X;\tht^{(t)})/\delta \tht$ denotes the derivative of the marginal log-likelihood with respect to the parameters $\tht$ evaluated at $\tht^{(t)}$. 
\end{itemize}
The T-step is based on an approximation to the conditional log-likelihood function, obtained by linearizing the marginal log-likelihood. 

In classification, $\set{Y}=\{1,\cdots , r\}$, the M-step is equivalent to finding $\tht$ so that the next equality holds:
\begin{equation}
\frac{\delta LL(X,Y;\tht)}{\delta \tht} = \frac{\delta LL(X,h(\cdot | X;\tht^{(t)}))}{\delta \tht},\nonumber
\end{equation}
with $LL(X,h(\cdot | X;\tht))=\sum_{\x \in X} \sum_{y \in \set{Y}} p_h(y|\x;\tht) \cdot \log h(\x,y; \tht)$. The TM differentiates the minimal sufficient statistics that depend on the class variable, $u$, from those independent of the class $v$, $s(X,Y)=(u(X,Y),v(X))$. For the particular case of generative classifiers from the exponential family, the TM iterates by updating the minimal sufficient statistics that depend on the class. Following our notation, it can be shown that in supervised classification TM reduces to:
$$\bd{u}^{(t+1)}= \bd{u}^{(t)} + u(X,Y) - u(X,h^{(t)}),$$
where the parameters of the model at iteration $t+1$ are given by the maximum likelihood sufficient statistics $\bd{s}=(\bd{u}^{(t+1)},\bd{v}^{(0)})$, with $\bd{v}^{(0)} = v(X)$. In summary, for the particular case of generative classifiers from the exponential family using ML, and a learning rate of $lr=1$, TM and RC are equivalent. 

The TM algorithm is a general method designed to maximize the conditional likelihood by using the parameters of the joint model. This algorithm has had very limited use, probably due to the difficulty of its implementation, stemming from its abstract mathematical definition, which offers very limited guidance on adapting the algorithm for practical use. For instance, one of the few examples of TM usage is outlined in \cite{Guzman07}, where the authors used it to learn the structure and parameters of classifiers based on Bayesian networks with categorical variables. The primary methodological challenge in that work involves updating the minimal sufficient statistics of conditional distributions over categorical variables while maintaining their consistency. This unnecessarily complicates the implementation of the learning method, particularly since its complexity strongly depends on the number of states of the categorical variables. This complexity becomes evident when compared to RC, which only requires iteratively updating the statistics using those derived from probabilistically labeled data.

\section{Experiments with RC for NB and QDA} \label{sec:experiments}


In this section, we summarize the experimental results obtained with NB and QDA, which were introduced in Section~\ref{sec:examples_generative} as illustrative examples of generative classifiers. We evaluate the performance of our proposed RC algorithm by comparing it with the maximum likelihood closed-form algorithm (ML) and gradient descent (GD) on the average log loss. Specifically, we analyze in this section the behavior of NB and QDA when trained using ML, GD, and RC (initialized with ML parameters). The results are provided in terms of empirical error on both training and test datasets. In~\ref{app:experiments_map}, we include additional results using the maximum a-posteriori closed-form algorithm (MAP) and RC with MAP. The implementations of the classifiers, learning algorithms, and experiments are available online in the public Python repository at \url{https://gitlab.bcamath.org/aperez/risk-based_calibration}.

\begin{table}[t]
\centering
\begin{tabular}{rlrr}
\toprule
index & data & m & n \\
\midrule
1 & iris & 150 & 4 \\
2 & sonar & 207 & 60 \\
3 & climate model & 360 & 18 \\
4 & indian liver & 579 & 10 \\
5 & vehicle & 846 & 18 \\
6 & german numer & 1000 & 24 \\
7 & splice & 1000 & 60 \\
8 & qsar & 1055 & 41 \\
9 & svmguide3 & 1243 & 21 \\
10 & redwine & 1599 & 11 \\
11 & optdigits & 5620 & 62 \\
12 & satellite & 6435 & 36 \\
13 & pulsar & 17898 & 8 \\
14 & magic & 19020 & 10 \\
15 & catsvsdogs & 23262 & 512 \\
16 & yearbook & 37921 & 512 \\
17 & adult & 48842 & 14 \\
18 & cifar10 & 60000 & 512 \\
19 & fashion mnist & 70000 & 512 \\
20 & mnist & 70000 & 512 \\
\bottomrule
\end{tabular}
\caption{Data sets. "Index" column contains the identifier used in the tables with empirical results, and the columns "$m$" and "$n$" correspond to the number of instances and variables of each dataset,respectively.}\label{table:data}
\end{table}

For the experiments, we used 20 public available datasets~\cite{Dua:2017, Elson07, Krizhevsky09, lecun2010mnist, tensorflow2015, Xiao17}, each characterized by a different number of instances ($m$) and variables ($n$).
Table \ref{table:data} provides an overview of the datasets used for the experiments, where the "Index" column serves as the identifier referenced along the empirical analysis. This comprehensive collection covers a wide range of domains and complexities, ensuring a robust evaluation of the proposed method in diverse real-world scenarios. Note that the 512 features of the datasets $\{15, 16, 18, 19, 20\}$ are provided by ResNet18~\cite{He16} pre-trained deep neural networks for image classification problems. These specific datasets provide deep feature representations for downstream classification containing a larger number of instances and input variables compared to the others. RC and GD algorithms have been run $64$ iterations with a fixed learning rate of $lr= 0.1$. 

Each experiment is repeated five times using random splits of the available data into training sets (75\%) and test sets (25\%). RC is compared with GD using the same initial parameter values. The main results are summarized in tables indexed by dataset, with the following column descriptions:

\begin{itemize}
    \item \textbf{ML} (MAP in~\ref{app:experiments_map}): This column displays the empirical error rate, expressed as a percentage of misclassified instances, obtained using ML parameters. These parameters served as the initialization for both RC and GD.
    \item \textbf{RC} and \textbf{GD}: These columns show the empirical error rates achieved by RC and GD, respectively, once the stopping condition is met. The best result for each dataset is highlighted in bold.
    \item \textbf{Iter}: This column contains the iteration at which RC and GD reached their minimum soft error. The values are averages over the five repetitions.
    \item \textbf{Reach}: This column indicates average the number of iterations required for RC to achieve a soft error less than or equal to the lowest soft error obtained by GD in train. This value is only shown when RC outperforms GD across the five repetitions. A smaller "Reach" value indicates that RC reduced errors more steeply and quickly than GD.
\end{itemize}
In addition, the "avg." row presents the average ranking position (1, 2 or 3) of the generative classifier, RC, and GD in terms of training and test errors across all datasets.

\subsection{RC for NB}
NB is a classification model that deals with discrete variables. To adapt the datasets from Table \ref{table:data} for use with NB models, each continuous variable is discretized into 5 categories according to a k-means strategy~\cite{dash2011discretization}, where the values in each bin belong to the same cluster. The experimental results for NB trained with ML, RC and GD initialized with ML parameters are detailed in Table~\ref{table:results_nb_ml}. A summary of key comparisons between these three methods is shown in Table~\ref{table:comparison_nb_ml}, which shows the number of datasets where RC performs better, equal to, or worse than ML and GD in terms of average training and test errors.

\begin{table*}[t!]
   \centering
    \begin{tabular}{l|cc|ccc|ccc|c}
    \hline
    \textbf{Index} & \multicolumn{2}{c|}{\textbf{ML}} & \multicolumn{3}{c|}{\textbf{RC}} & \multicolumn{3}{c|}{\textbf{GD}} \\
    & Training (\%) & Test (\%) & Training (\%) & Test (\%) & Iter & Training (\%) & Test (\%) & Iter & Reach \\
    \hline
    1 & $04 \pm 01$ & $05 \pm 03$ & $\textbf{03} \pm 01$ & $\textbf{04} \pm 03$ & 41.0 & $04 \pm 01$ & $06 \pm 04$ & 64.0 & 06.2 \\
2 & $12 \pm 00$ & $28 \pm 05$ & $\textbf{00} \pm 01$ & $\textbf{03} \pm 04$ & 38.2 & $03 \pm 01$ & $14 \pm 06$ & 53.8 & 06.8 \\
3 & $03 \pm 00$ & $08 \pm 01$ & $\textbf{02} \pm 00$ & $\textbf{03} \pm 02$ & 40.2 & $02 \pm 00$ & $07 \pm 01$ & 64.0 & - \\
4 & $30 \pm 01$ & $36 \pm 03$ & $30 \pm 01$ & $36 \pm 03$ & 01.0 & $\textbf{29} \pm 01$ & $\textbf{36} \pm 03$ & 02.8 & - \\
5 & $35 \pm 01$ & $38 \pm 03$ & $35 \pm 01$ & $38 \pm 03$ & 01.0 & $\textbf{34} \pm 00$ & $\textbf{38} \pm 03$ & 09.8 & - \\
6 & $22 \pm 00$ & $25 \pm 02$ & $22 \pm 00$ & $24 \pm 02$ & 01.2 & $\textbf{21} \pm 00$ & $\textbf{22} \pm 00$ & 51.2 & - \\
7 & $08 \pm 00$ & $10 \pm 02$ & $\textbf{02} \pm 00$ & $\textbf{02} \pm 01$ & 64.0 & $05 \pm 00$ & $07 \pm 02$ & 64.0 & 07.2 \\
8 & $\textbf{17} \pm 00$ & $20 \pm 00$ & $\textbf{17} \pm 00$ & $20 \pm 00$ & 01.0 & $17 \pm 00$ & $\textbf{20} \pm 00$ & 01.2 & 01.0 \\
9 & $22 \pm 00$ & $24 \pm 02$ & $\textbf{15} \pm 00$ & $\textbf{16} \pm 01$ & 29.2 & $21 \pm 01$ & $23 \pm 02$ & 01.6 & 01.6 \\
10 & $\textbf{41} \pm 00$ & $44 \pm 01$ & $\textbf{41} \pm 00$ & $44 \pm 01$ & 01.0 & $41 \pm 00$ & $\textbf{43} \pm 01$ & 03.2 & - \\
11 & $06 \pm 00$ & $08 \pm 00$ & $\textbf{05} \pm 00$ & $\textbf{06} \pm 00$ & 07.4 & $06 \pm 00$ & $08 \pm 00$ & 01.0 & 01.0 \\
12 & $20 \pm 00$ & $20 \pm 00$ & $\textbf{20} \pm 00$ & $\textbf{20} \pm 01$ & 02.2 & $20 \pm 00$ & $20 \pm 00$ & 01.0 & 01.0 \\
13 & $03 \pm 00$ & $03 \pm 00$ & $\textbf{02} \pm 00$ & $\textbf{02} \pm 00$ & 09.6 & $03 \pm 00$ & $03 \pm 00$ & 09.2 & 02.0 \\
14 & $24 \pm 00$ & $24 \pm 00$ & $\textbf{17} \pm 00$ & $\textbf{17} \pm 00$ & 64.0 & $18 \pm 00$ & $18 \pm 00$ & 64.0 & 10.2 \\
15 & $03 \pm 00$ & $03 \pm 00$ & $\textbf{01} \pm 00$ & $\textbf{01} \pm 00$ & 63.2 & $03 \pm 00$ & $03 \pm 00$ & 01.0 & 01.0 \\
16 & $15 \pm 00$ & $14 \pm 00$ & $\textbf{07} \pm 00$ & $\textbf{07} \pm 00$ & 64.0 & $13 \pm 00$ & $13 \pm 00$ & 41.4 & 06.0 \\
17 & $18 \pm 00$ & $18 \pm 00$ & $\textbf{15} \pm 00$ & $\textbf{15} \pm 00$ & 09.0 & $18 \pm 00$ & $18 \pm 00$ & 02.4 & 02.0 \\
18 & $21 \pm 00$ & $21 \pm 00$ & $\textbf{11} \pm 00$ & $\textbf{11} \pm 00$ & 57.4 & $20 \pm 00$ & $21 \pm 00$ & 19.2 & 02.0 \\
19 & $20 \pm 00$ & $20 \pm 00$ & $\textbf{12} \pm 00$ & $\textbf{13} \pm 00$ & 33.2 & $20 \pm 00$ & $20 \pm 00$ & 07.2 & 01.4 \\
20 & $11 \pm 00$ & $12 \pm 00$ & $\textbf{04} \pm 00$ & $\textbf{04} \pm 00$ & 60.6 & $11 \pm 00$ & $12 \pm 00$ & 11.4 & 01.8 \\
\hline
\textbf{avg.} & 2.83 & 2.77 & 1.20 & 1.32 & - & 1.98 & 1.90 & - & - \\
\hline
    \end{tabular}
    \caption{Comparison of NB in terms of training and test errors learned with ML, RC, and GD.}
    \label{table:results_nb_ml}
\end{table*}

\begin{table*}[t]
    \centering
    \begin{tabular}{l|ccc|ccc}
    \hline
    & \multicolumn{3}{c|}{\textbf{RC vs ML}} & \multicolumn{3}{c}{\textbf{RC vs GD}} \\
    & Better & Equal & Worse & Better & Equal & Worse \\
    \hline
    Training & 18 & 2 & 0 & 17 & 0 & 3 \\
    Test & 17 & 3 & 0 & 15 & 0 & 5 \\
    \hline
    \end{tabular}
    \caption{Summary of the comparisons of RC against ML and GD in terms of the training and test errors across all datasets.}
    \label{table:comparison_nb_ml}
\end{table*}

Overall, Table~\ref{table:results_nb_ml} shows that RC consistently outperforms both ML and GD, achieving lower training and test errors, as well as superior average ranking. Notably, RC is also more efficient according to the "Reach" column since RC achieves a lower error than GD, across the five repetitions, in 15 out of 20 datasets while requiring fewer iterations to converge in all these cases.  Its ability to significantly reduce errors in challenging datasets (e.g., datasets 2, 14, 16, 18) further demonstrates its effectiveness. Also note that RC maintains low standard deviations similar to ML, indicating stable and reliable results across multiple runs. These findings align with the results obtained using Maximum A Posteriori (MAP) estimation (see~\ref{app:experiments_map} for details).

Table~\ref{table:comparison_nb_ml} clearly shows RC's superior performance. When compared to ML, RC achieves better results in 18 out of 20 datasets for training errors and 17 out of 20 for test errors. In comparison with GD, RC's advantage is also clear, outperforming GD in 17 datasets for training errors and 15 for test errors. These results strongly support the aforementioned observations about the performance of RC, showing that it is consistently better across the majority of datasets, both in training and testing scenarios.

\subsection{RC for QDA}
\begin{table*}[t!]
    \centering
    \begin{tabular}{l|cc|ccc|ccc|c}
    \hline
    \textbf{Index} & \multicolumn{2}{c|}{\textbf{ML}} & \multicolumn{3}{c|}{\textbf{RC}} & \multicolumn{3}{c|}{\textbf{GD}} \\
    & Training (\%) & Test (\%)& Training (\%)& Test (\%)& Iter & Training (\%)& Test (\%)& Iter & Reach \\
    \hline
    1 & $02 \pm 00$ & $\textbf{01} \pm 01$ & $\textbf{01} \pm 00$ & $\textbf{01} \pm 01$ & 64.0 & $02 \pm 00$ & $\textbf{01} \pm 01$ & 54.6 & 01.0 \\
2 & $\textbf{00} \pm 00$ & $\textbf{00} \pm 00$ & $\textbf{00} \pm 00$ & $\textbf{00} \pm 00$ & 64.0 & $\textbf{00} \pm 00$ & $\textbf{00} \pm 00$ & 01.0 & 01.0 \\
3 & $01 \pm 00$ & $01 \pm 00$ & $\textbf{00} \pm 00$ & $\textbf{00} \pm 00$ & 64.0 & $\textbf{00} \pm 00$ & $\textbf{00} \pm 00$ & 64.0 & 04.0 \\
4 & $44 \pm 01$ & $44 \pm 04$ & $\textbf{32} \pm 01$ & $\textbf{33} \pm 04$ & 06.0 & $44 \pm 01$ & $44 \pm 04$ & 01.0 & 01.0 \\
5 & $08 \pm 00$ & $09 \pm 01$ & $\textbf{02} \pm 00$ & $\textbf{03} \pm 00$ & 64.0 & $08 \pm 00$ & $09 \pm 01$ & 01.0 & 01.0 \\
6 & $20 \pm 00$ & $22 \pm 01$ & $\textbf{11} \pm 00$ & $\textbf{11} \pm 01$ & 64.0 & $20 \pm 00$ & $22 \pm 01$ & 01.0 & 01.0 \\
7 & $02 \pm 00$ & $02 \pm 00$ & $\textbf{00} \pm 00$ & $\textbf{00} \pm 00$ & 64.0 & $02 \pm 00$ & $02 \pm 00$ & 01.0 & 01.0 \\
8 & $20 \pm 00$ & $19 \pm 01$ & $\textbf{02} \pm 00$ & $\textbf{03} \pm 00$ & 64.0 & $20 \pm 00$ & $19 \pm 01$ & 01.0 & 01.0 \\
9 & $16 \pm 00$ & $16 \pm 01$ & $\textbf{13} \pm 00$ & $\textbf{14} \pm 01$ & 09.2 & $16 \pm 00$ & $16 \pm 01$ & 01.0 & 01.0 \\
10 & $58 \pm 01$ & $58 \pm 03$ & $58 \pm 01$ & $58 \pm 03$ & 02.0 & $\textbf{45} \pm 00$ & $\textbf{47} \pm 02$ & 02.0 & - \\
11 & $01 \pm 00$ & $01 \pm 00$ & $\textbf{00} \pm 00$ & $\textbf{00} \pm 00$ & 64.0 & $01 \pm 00$ & $01 \pm 00$ & 01.8 & 01.0 \\
12 & $11 \pm 00$ & $11 \pm 00$ & $\textbf{03} \pm 00$ & $\textbf{03} \pm 00$ & 64.0 & $11 \pm 00$ & $11 \pm 00$ & 01.0 & 01.0 \\
13 & $03 \pm 00$ & $03 \pm 00$ & $\textbf{02} \pm 00$ & $\textbf{02} \pm 00$ & 06.8 & $02 \pm 00$ & $02 \pm 00$ & 34.8 & 02.0 \\
14 & $21 \pm 00$ & $21 \pm 00$ & $\textbf{17} \pm 00$ & $\textbf{17} \pm 00$ & 12.2 & $21 \pm 00$ & $21 \pm 00$ & 02.0 & 01.8 \\
15 & $00 \pm 00$ & $00 \pm 00$ & $\textbf{00} \pm 00$ & $\textbf{00} \pm 00$ & 29.4 & $00 \pm 00$ & $00 \pm 00$ & 01.0 & 01.0 \\
16 & $09 \pm 00$ & $09 \pm 00$ & $\textbf{00} \pm 00$ & $\textbf{00} \pm 00$ & 64.0 & $09 \pm 00$ & $09 \pm 00$ & 01.0 & 01.0 \\
17 & $19 \pm 00$ & $19 \pm 00$ & $\textbf{16} \pm 00$ & $\textbf{16} \pm 00$ & 07.6 & $19 \pm 00$ & $19 \pm 00$ & 01.0 & 01.0 \\
18 & $06 \pm 00$ & $06 \pm 00$ & $\textbf{00} \pm 00$ & $\textbf{00} \pm 00$ & 64.0 & $06 \pm 00$ & $06 \pm 00$ & 01.0 & 01.0 \\
19 & $08 \pm 00$ & $08 \pm 00$ & $\textbf{00} \pm 00$ & $\textbf{00} \pm 00$ & 64.0 & $08 \pm 00$ & $08 \pm 00$ & 01.0 & 01.0 \\
20 & $01 \pm 00$ & $01 \pm 00$ & $\textbf{00} \pm 00$ & $\textbf{00} \pm 00$ & 64.0 & $01 \pm 00$ & $01 \pm 00$ & 01.0 & 01.0 \\
\hline
\textbf{avg.} & 2.58 & 2.58 & 1.12 & 1.18 & - & 2.30 & 2.25 & - & - \\
\hline
    \end{tabular}
    \caption{Comparison of QDA in terms of error in training and test sets trained with ML, RC, and GD.}
    \label{table:results_qda_ml}
\end{table*}

\begin{table*}[t]
    \centering
    \begin{tabular}{l|ccc|ccc}
    \hline
    & \multicolumn{3}{c|}{\textbf{RC vs ML}} & \multicolumn{3}{c}{\textbf{RC vs GD}} \\
    & Better & Equal & Worse & Better & Equal & Worse \\
    \hline
    Training & 19 & 1 & 0 & 17 & 2 & 1 \\
    Test & 18 & 2 & 0 & 16 & 3 & 1 \\
    \hline
    \end{tabular}
    \caption{Summary of the comparisons of RC against ML and GD in terms of the training and test errors across all datasets.}
    \label{table:comparison_qda_ml}
\end{table*}

The results for learning QDA using ML, RC based on ML, and GD initialized with ML are presented in Table~\ref{table:results_qda_ml}. A summary of the comparisons between ML, RC, and GD is shown in Table~\ref{table:comparison_qda_ml}. 
Overall, RC clearly outperforms ML and GD in terms of training and test errors, as well as in average rankings demonstrating its robustness when applied to QDA. Notably, according to the "Reach" column, RC achieves a lower error than GD in 19 out of 20 datasets across the five repetitions, while requiring fewer iterations to converge in all these cases. As for NB, when  RC is applied to QDA, it maintains low standard deviations similar to ML and GD, indicating stable and reliable results across multiple runs, while offering improved accuracy. In many cases, RC conducts a higher number of iterations (often using the maximum of 64 iterations) resulting in substantially improved performance. For example, RC achieves the most dramatic error reductions in dataset 8 (from 20\% to 2\% in training, and from 19\% to 3\% in test), and achieves perfect classification in 7 datasets. Moreover, RC shows remarkable improvement in datasets 4, 5, 6 and 12 compared to both ML and GD.

In Table~\ref{table:comparison_qda_ml}, we can see that RC achieves better results in 19 out of 20 datasets for training errors and 18 out of 20 for test errors. In comparison with GD, the advantage of RC is also evident, outperforming GD in 17 datasets for training errors and 16 for test errors. There is only one case (dataset 10) where GD performs better in both training and test errors. These results strongly support the effectiveness of RC for QDA, showing that its improved performance is consistent across the vast majority of datasets, both in training and testing scenarios. The ability of RC to significantly reduce error rates, especially in challenging datasets, underscores its potential as a robust and efficient technique to minimize the classification error.

\section{Conclusions} \label{sec:conclusion}
This work introduces \textit{Risk-based Calibration (RC)}, an iterative learning algorithm designed to minimize the empirical error of generative classifiers. Unlike traditional methods that rely on likelihood scores unrelated to classification performance, RC directly calibrates data statistics using a soft 0-1 loss, explicitly targeting training error minimization. This approach yields valid parameter estimates without requiring additional transformations or complex constraint-handling techniques, addressing limitations commonly encountered in gradient descent-based methods.

RC is applicable to any generative classifier that follows a two-step closed-form learning procedure: (i) collecting statistics from the data, and (ii) computing parameters analytically from those statistics. This includes widely used models from the exponential family, such as \textit{naïve Bayes} and \textit{quadratic discriminant analysis}, which are used in this work to showcase the benefits of RC.

Comprehensive experiments on 20 diverse datasets—ranging from 150 to 70,000 instances and from 4 to 512 features—demonstrate that RC consistently achieves lower training and test errors compared to both standard closed-form methods and gradient-based alternatives.

In summary, RC offers a practical and effective solution for learning generative classifiers, combining computational efficiency, ease of implementation, and improved classification performance. The full implementation, including classifiers, learning algorithms, and experimental setups, is publicly available at \url{https://gitlab.bcamath.org/aperez/risk-based_calibration}.

\section*{Acknowledgements}

This work has been carried out with financial support from:

\begin{itemize}
  \item The Basque Government through the BERC 2022--2025 program and Elkartek program (SONETO,\\ KK-2023/00038).
  \item The Ministry of Science, Innovation and Universities, under the BCAM Severo Ochoa accreditation \\\texttt{CEX2021-001142-S/MICIN/AEI/10.13039/501100011033}.
\end{itemize}

\newpage

\appendix
\renewcommand{\thesection}{Appendix \Alph{section}}

\section{Organization}
This supplementary material provides additional details related to the experiments. In \ref{app:GD}, we present a full description of the gradient descent updating rule used to learn NB and QDA in the conducted experiments. \ref{app:MAP} outlines the maximum a posteriori (MAP) Bayesian estimation of the parameters for discrete naïve Bayes (NB) and quadratic discriminant analysis (QDA). \ref{app:experiments_map} contains the empirical analysis of RC using MAP for learning NB and QDA. Finally, \ref{app:DFE} presents the empirical comparison between RC and Discriminative Frequency Estimate~\cite{su2008discriminative}.

\section{Gradient descent}\label{app:GD}

In this section, we first describe gradient descent for the empirical risk under the log loss as a method for updating the parameters of the models. These parameters are subject to specific constraints: in NB, they represent probability distributions for categorical variables and must be non-negative and sum to one, while in QDA, some parameters form covariance matrices that must be positive semidefinite. A common approach to handle these constraints is to update the parameters using gradient descent and then project them onto the feasible set. This method, known as projected gradient descent (see \cite{bertsekas1997nonlinear}), ensures that the parameter constraints are maintained throughout the optimization process. However, this approach comes at the cost of additional computational complexity, as the projection step can be computationally expensive, especially when dealing with complex constraints or high-dimensional data.

All the classifiers have been expressed into the exponential family form, $$p_h(y|\x) \propto \e{\bd{\eta}^T\cdot s(\x,y) + A_y(\bd{\eta}^T)}.$$ This simplifies obtaining a clear expression for the gradient of the parameters and puts all the considered models into a comparable form.

The exponential family form of the conditional probability modeled by NB is given by feature mapping corresponding to the one-hot encoding of each feature and the class
$\phi(\x,y)=(\1{1=y},\1{1=y}\1{1=x_1},...,\1{1=y}\1{r_1=x_1},...,,\1{1=y},\1{1=y}\1{1=x_n},...,\1{r=y}\1{r_n=x_n},...,\1{r=y}\1{1=x_n},...,\1{r=y}\1{r_n=x_n})$ and the parameters
$\bd{\eta}=(\bd{\eta_0},\bd{\eta}_{1,0},...,\bd{\eta}_{n,0},... ,\bd{\eta}_{1,r},...,\bd{\eta}_{n,r})$ with $\bd{\eta}_0=(\eta_{0,y}= \log p(y))_{y=1}^r$ and $\bd{\eta}_{i,y}=(\eta_{i,x_i,y} = \log p(x_i|y))_{x_i=1}^{r_i}$ for $i=1,...,n$ and $y \in \set{Y}$. For NB $A_y(\bd{\eta}^T)=0$. The gradient descent updating rule is given by:
\begin{align}
\eta_{0,y'}=& \eta_{0,y'} - \frac{lr}{m} \sum_{\x,y \in X,Y}  (p_h(y'|\x)- \1{y'=y})\nonumber\\
\eta_{i,x_i',y'}=& \eta_{i,x_i',y'}\nonumber\\ &- \frac{lr}{m} \sum_{\x,y \in X,Y} \1{x_i'= x_i} \cdot(p_h(y'|\x) - \1{y'=y})\nonumber
\end{align}
for $y'=1,...,r$, $i=1,...,n$, and $x'_i=1,...,r_i$.

Then, after applying the gradient descent updating rule, the natural parameters are transformed into probabilities by exponentiation and by projecting them into the simplex. Alternatives to the projection to the simplex include using softmax to obtain proper probability distributions. Unfortunately, by using transformations, the descent in the average negative log loss is no longer guaranteed.

The exponential-family distribution form of the conditional probability modeled by QDA is given by the feature mapping corresponding to the class one-hot coding $\phi(\x,y)=[\1{1=y},\1{1=y}\cdot \x,\1{1=y} \x \cdot \x^T,..., \1{r=y},\1{r=y}\cdot \x,\1{r=y}\cdot \x \cdot \x^T]$, where $[\cdots]$ rearranges its arguments to conform to a column vector; the parameters $\bd{\eta}=[\bd{\eta}_0,\bd{\eta}_1,\bd{\eta}_2]$ with $\bd{\eta}_0= (\eta_{0,y}=\log p(y))_{y=1}^r$, $\bd{\eta}_1= (\eta_{1,y}= \Sigma_y^{-1} \cdot \mu_y)_{y=1}^r$ and $\bd{\eta}_2=(\eta_{2,y}= -1/2 \cdot \Sigma_y^{-1})_{y=1}^r$, being $\mu_y$ and $\Sigma_y$ the mean vector and covariance matrix conditioned to $y$; and $A_y(\bd{\eta})= \frac{1}{4}\cdot \bd{\eta}_{2,y}^{-1}\cdot \bd{\eta}_{1,y}\cdot \bd{\eta}_{1,y}^T \cdot \bd{\eta}_{2,y}^{-1} - \frac{1}{2}\cdot \bd{\eta}_{2,y}^{-1}$. The gradient descent updating rules are given by:
\begin{align}
\bd{\eta}_{1,y'} =& \bd{\eta}_{1,y'} - \frac{lr}{m} \cdot \sum_{x,y} (p_h(y'|\x) - \1{y=y'})\nonumber\\
& \cdot(x + 1/2 \cdot \bd{\eta}_{2,y'}^{-1} \cdot \bd{\eta}_{1,y'})\nonumber\\
\bd{\eta}_{2,y'} =& \bd{\eta}_{2,y'} - \frac{lr}{m} \cdot \sum_{\x,y \in X,Y} (p_h(y'|\x) - \1{y=y'})\nonumber\\
&\cdot(\x^2 - 1/4 \cdot \bd{\eta}_{2,y'}^{-1}\cdot \bd{\eta}_{1,y'} \cdot \bd{\eta}_{1,y'}^T \cdot \bd{\eta}_{2,y'}^{-1} + 1/2\cdot tr(\bd{\eta}_{2,y'}^{-1}))\nonumber
\end{align}
for $y'=1,...,r$, and being $tr(\cdot)$ the trace of a matrix. 

Then, after every update of the gradient descent, the natural parameters $\bd{\eta}_0$ are transformed into probabilities by exponentiation and by projecting them into the simplex; $\bd{\eta}_{2,y}$ is transformed into the covariance matrix $\Sigma_y=-\frac{1}{2}\cdot \bd{\eta}_{2,y}^{-1}$ and is ensured to be a positive semi-definite matrix by: i) obtaining the singular value decomposition, ii) guaranteeing that all the eigenvalues are no smaller than $\epsilon= 10^{-2}$, and iii) reconstructing the matrix using the constrained eigenvalues. Again, by transforming the obtained parameters to fulfill their associated constraints (probabilities and covariance matrices) we cannot ensure that the average negative log loss descends.

\section{Maximum a posteriori parameter mapping}\label{app:MAP}

\noindent An alternative to the maximum likelihood parameter estimation is a Bayesian estimation of the parameters. In Bayesian estimation, we assume a prior for the distribution of the parameters. Then, given the data, we obtain the posterior distribution of the parameters and select the parameters of its mode (maximum a posteriori parameters, MAP). For certain parametric distributions, there are prior distributions over their parameters that allow obtaining the MAP parameters in closed form.

NB is based on the categorical distribution, and the Bayesian conjugate of the categorical distribution is the Dirichlet distribution. Let's take the categorical distribution $p(y)$ for $y \in \set{Y}=\{1,...,r\}$ and the prior Dirichlet distribution for its parameters
\begin{align}
\tht=(\theta_1,...,\theta_r) \sim Dir(\bd{\alpha}),\nonumber
\end{align}
with hyperparameters $\bd{\alpha}=(\alpha_1,...,\alpha_r)$. The posterior distribution of the parameters of $p(y)$ after observing $Y=\{y_i\}_{i=1}^m$ is given by:
\begin{align}
\tht \sim Dir(\bd{\alpha} + (m_1,...,m_r)),\nonumber
\end{align}
where $m_{y'}= \sum_{y \in Y} \1{y'=y}$ for $y' \in \set{Y}$. The MAP corresponds to:
\begin{equation}
\tht'_y= \frac{m_y + \alpha_y - 1}{m + \alpha - r},\nonumber    
\end{equation}
with $\alpha= \sum_{y=1}^r \alpha_y$. By taking $\alpha_y= m_0/r +1$ for $y \in 1$ we have the more intuitive MAP
\begin{equation}
\tht'_y= \frac{m_y + m_0/r}{m + m_0},\nonumber  
\end{equation}
where $m_0$ can be interpreted as the equivalent sample size of the prior. In the experiments with MAP we have taken $m_0= r$. The same analysis follows for all the conditional distributions that are involved in NB, $p(x_i|y)$ for $y \in \set{Y}$, and $i=1,...,n$ with $x_i \in \{1,...,r_i\}$.

\begin{table*}[t!]
    \centering
    \begin{tabular}{l|cc|ccc|ccc|c}
    \hline
    \textbf{Index} & \multicolumn{2}{c|}{\textbf{MAP}} & \multicolumn{3}{c|}{\textbf{RC}} & \multicolumn{3}{c|}{\textbf{GD}} \\
    & Training (\%)& Test (\%)& Training (\%)& Test (\%)& Iter & Training (\%)& Test (\%)& Iter & Reach \\
    \hline
    1 & $04 \pm 01$ & $06 \pm 03$ & $\textbf{04} \pm 01$ & $\textbf{04} \pm 03$ & 39.6 & $04 \pm 01$ & $04 \pm 03$ & 64.0 & 06.2 \\
2 & $13 \pm 00$ & $28 \pm 05$ & $\textbf{00} \pm 01$ & $\textbf{01} \pm 02$ & 55.0 & $03 \pm 01$ & $15 \pm 06$ & 56.0 & 07.0 \\
3 & $03 \pm 00$ & $08 \pm 01$ & $\textbf{02} \pm 00$ & $\textbf{02} \pm 00$ & 64.0 & $02 \pm 00$ & $07 \pm 01$ & 64.0 & 04.0 \\
4 & $30 \pm 01$ & $\textbf{35} \pm 03$ & $30 \pm 01$ & $\textbf{35} \pm 03$ & 01.0 & $\textbf{29} \pm 01$ & $\textbf{35} \pm 03$ & 03.4 & - \\
5 & $36 \pm 01$ & $39 \pm 03$ & $36 \pm 01$ & $39 \pm 03$ & 01.0 & $\textbf{34} \pm 00$ & $\textbf{37} \pm 03$ & 21.0 & - \\
6 & $22 \pm 00$ & $25 \pm 01$ & $22 \pm 00$ & $25 \pm 01$ & 01.0 & $\textbf{21} \pm 00$ & $\textbf{21} \pm 00$ & 57.0 & - \\
7 & $08 \pm 00$ & $10 \pm 02$ & $\textbf{02} \pm 00$ & $\textbf{02} \pm 01$ & 64.0 & $05 \pm 00$ & $07 \pm 02$ & 64.0 & 07.2 \\
8 & $\textbf{18} \pm 00$ & $\textbf{20} \pm 00$ & $\textbf{18} \pm 00$ & $20 \pm 00$ & 01.0 & $18 \pm 00$ & $20 \pm 01$ & 02.2 & - \\
9 & $22 \pm 01$ & $24 \pm 02$ & $\textbf{15} \pm 00$ & $\textbf{16} \pm 00$ & 29.4 & $21 \pm 00$ & $22 \pm 02$ & 02.0 & 02.0 \\
10 & $41 \pm 00$ & $\textbf{43} \pm 01$ & $41 \pm 00$ & $\textbf{43} \pm 01$ & 01.0 & $\textbf{41} \pm 00$ & $43 \pm 01$ & 03.4 & - \\
11 & $07 \pm 00$ & $07 \pm 00$ & $\textbf{06} \pm 00$ & $\textbf{06} \pm 00$ & 06.4 & $07 \pm 00$ & $07 \pm 00$ & 09.8 & 01.6 \\
12 & $20 \pm 00$ & $20 \pm 00$ & $20 \pm 00$ & $20 \pm 00$ & 02.2 & $\textbf{20} \pm 00$ & $\textbf{20} \pm 00$ & 16.0 & - \\
13 & $03 \pm 00$ & $03 \pm 00$ & $\textbf{02} \pm 00$ & $\textbf{02} \pm 00$ & 09.6 & $03 \pm 00$ & $03 \pm 00$ & 09.4 & 02.0 \\
14 & $24 \pm 00$ & $24 \pm 00$ & $\textbf{17} \pm 00$ & $\textbf{17} \pm 00$ & 64.0 & $18 \pm 00$ & $18 \pm 00$ & 64.0 & 10.2 \\
15 & $03 \pm 00$ & $03 \pm 00$ & $\textbf{01} \pm 00$ & $\textbf{01} \pm 00$ & 64.0 & $03 \pm 00$ & $03 \pm 00$ & 01.0 & 01.0 \\
16 & $15 \pm 00$ & $14 \pm 00$ & $\textbf{07} \pm 00$ & $\textbf{07} \pm 00$ & 64.0 & $13 \pm 00$ & $13 \pm 00$ & 41.6 & 06.0 \\
17 & $18 \pm 00$ & $18 \pm 00$ & $\textbf{15} \pm 00$ & $\textbf{15} \pm 00$ & 09.0 & $18 \pm 00$ & $18 \pm 00$ & 02.4 & 02.0 \\
18 & $21 \pm 00$ & $21 \pm 00$ & $\textbf{12} \pm 00$ & $\textbf{12} \pm 00$ & 45.8 & $20 \pm 00$ & $21 \pm 00$ & 19.4 & 02.0 \\
19 & $20 \pm 00$ & $20 \pm 00$ & $\textbf{13} \pm 00$ & $\textbf{13} \pm 00$ & 28.2 & $20 \pm 00$ & $20 \pm 00$ & 17.4 & 02.0 \\
20 & $11 \pm 00$ & $12 \pm 00$ & $\textbf{06} \pm 00$ & $\textbf{06} \pm 00$ & 41.4 & $11 \pm 00$ & $12 \pm 00$ & 16.4 & 02.0 \\
\hline
\textbf{avg.} & 2.80 & 2.67 & 1.38 & 1.32 & - & 1.82 & 2.00 & - & - \\
\hline
    \end{tabular}
    \caption{Comparison of NB in terms of error in training and test sets trained with MAP, RC, and GD.}
    \label{table:results_nb_map}
\end{table*}

\begin{table*}[t]
    \centering
    \begin{tabular}{l|ccc|ccc}
    \hline
    & \multicolumn{3}{c|}{\textbf{RC vs MAP}} & \multicolumn{3}{c}{\textbf{RC vs GD}} \\
    & Better & Equal & Worse & Better & Equal & Worse \\
    \hline
    Training & 15 & 5 & 0 & 15 & 0 & 5 \\
    Test & 15 & 4 & 1 & 16 & 1 & 3 \\
    \hline
    \end{tabular}
    \caption{Summary of the comparisons of RC against MAP and GD in terms of the average training and test errors across all datasets.}
    \label{table:comparison_nb_map}
\end{table*}

QDA is based on a categorical distribution $p(y)$ and $n$-dimensional Gaussian density functions $p(\x|y)$ for $y \in \set{Y}$. The Bayesian conjugate of the parameters of a multivariate Gaussian distribution is the normal distribution for the mean and the inverse-Wishart distribution for the covariance:
\begin{align}
\mu|\Sigma \sim& \set{N}(\mu_0,1/\kappa_0\Sigma)\nonumber\\
\Sigma \sim& \set{W}^{-1}(T_0, \nu_0)\nonumber
\end{align}
The posterior distribution for the parameters given the observations $X=\{\x_1,...,\x_m\}$ is given by
\begin{align}
\mu|\Sigma \sim \set{N}(\mu', 1/\kappa'\Sigma)\nonumber\\
\Sigma \sim \set{W}^{-1}(\Sigma',\nu')\nonumber
\end{align}
Then, the MAP parameters for the mean and the covariance matrix are
\begin{eqnarray}
\mu'&=& \frac{\kappa_0\mu_0 + m\hat{\mu}}{\kappa_0 + m}\nonumber\\ 
\Sigma'&=& \frac{T_0 + m\hat{\Sigma} + \frac{\kappa_0 \cdot m}{\kappa_0 + m}(\hat{\mu}-\mu_0)\cdot (\hat{\mu}-\mu_0)^T}{\nu_0+m+n+1},\nonumber
\end{eqnarray}
where $\hat{\mu}$ and $\hat{\Sigma}$ are the sample mean and covariance matrix. We propose the following re-parametrization in terms of $m_1, m_2 \geq 0$, $\kappa_0= m_1$, $T_0= m_2\Sigma_0$ and $\nu_0=(m_2- n -1)$, for the sake of interpretability. Under this re-parametrization and neglecting the term $(\kappa_0 \cdot m)(\hat{\mu}-\mu_0)\cdot (\hat{\mu}-\mu_0)^T/((\nu_0+m+n+1)\cdot(\kappa_0 + m))$ because usually $\kappa_0<<m$, we have the next intuitive MAP parameters in terms of prior mean vector $\mu_0$ and covariance matrix $\Sigma_0$ with weights $m_1$ and $m_2$, respectively:
\begin{align}
\mu'= \frac{m_1\cdot \mu_0 + m\hat{\mu}}{m_1 + m}\nonumber\\ 
\Sigma'= \frac{m_2\cdot \Sigma_0 + m\hat{\Sigma}}{m_2 + m}\nonumber
\end{align}
Here, $m_1$ and $m_2$ can be interpreted as the equivalent sample size of the priors for the mean vector and covariance matrix, respectively. In the experiments with MAP we have taken $m_1=m_2=10$.
This MAP estimate is used for the parameters of $p(\x|y)$ for $y \in \set{Y}$, while the MAP parameters of $p(y)$ are obtained using the same procedure of NB. We have followed a similar approach for the parameter mapping used in the closed-form algorithm of LR.

\section{Empirical results with MAP}\label{app:experiments_map}

\begin{table*}[t]
    \centering
    \begin{tabular}{l|cc|ccc|ccc|c}
    \hline
    \textbf{Index} & \multicolumn{2}{c|}{\textbf{MAP}} & \multicolumn{3}{c|}{\textbf{RC}} & \multicolumn{3}{c|}{\textbf{GD}} \\
    & Training (\%)& Test (\%)& Training (\%)& Test (\%)& Iter & Training (\%) & Test (\%) & Iter & Reach \\
    \hline
    1 & $\textbf{03} \pm 00$ & $\textbf{02} \pm 01$ & $\textbf{03} \pm 00$ & $\textbf{02} \pm 01$ & 01.0 & $\textbf{03} \pm 00$ & $\textbf{02} \pm 01$ & 02.0 & 01.0 \\
2 & $\textbf{00} \pm 00$ & $\textbf{00} \pm 00$ & $\textbf{00} \pm 00$ & $\textbf{00} \pm 00$ & 01.0 & $\textbf{00} \pm 00$ & $\textbf{00} \pm 00$ & 01.0 & 01.0 \\
3 & $01 \pm 00$ & $\textbf{01} \pm 00$ & $01 \pm 00$ & $\textbf{01} \pm 00$ & 01.0 & $\textbf{01} \pm 00$ & $02 \pm 01$ & 01.6 & - \\
4 & $48 \pm 01$ & $47 \pm 04$ & $\textbf{25} \pm 00$ & $\textbf{24} \pm 01$ & 32.0 & $48 \pm 01$ & $47 \pm 04$ & 01.0 & 01.0 \\
5 & $\textbf{10} \pm 00$ & $\textbf{12} \pm 02$ & $10 \pm 00$ & $\textbf{12} \pm 02$ & 01.0 & $10 \pm 00$ & $\textbf{12} \pm 02$ & 01.0 & 01.0 \\
6 & $20 \pm 00$ & $23 \pm 01$ & $\textbf{11} \pm 00$ & $\textbf{11} \pm 01$ & 64.0 & $20 \pm 00$ & $23 \pm 01$ & 01.0 & 01.0 \\
7 & $02 \pm 00$ & $03 \pm 00$ & $\textbf{00} \pm 00$ & $\textbf{00} \pm 00$ & 64.0 & $02 \pm 00$ & $03 \pm 00$ & 01.0 & 01.0 \\
8 & $\textbf{22} \pm 00$ & $\textbf{21} \pm 01$ & $22 \pm 00$ & $\textbf{21} \pm 01$ & 01.0 & $22 \pm 00$ & $\textbf{21} \pm 01$ & 01.0 & 01.0 \\
9 & $\textbf{16} \pm 00$ & $\textbf{16} \pm 01$ & $\textbf{16} \pm 00$ & $16 \pm 01$ & 01.0 & $\textbf{16} \pm 00$ & $16 \pm 01$ & 01.0 & 01.0 \\
10 & $40 \pm 01$ & $42 \pm 03$ & $\textbf{36} \pm 00$ & $\textbf{37} \pm 02$ & 64.0 & $40 \pm 01$ & $42 \pm 03$ & 01.0 & 01.0 \\
11 & $01 \pm 00$ & $01 \pm 00$ & $\textbf{00} \pm 00$ & $\textbf{00} \pm 00$ & 64.0 & $01 \pm 00$ & $01 \pm 00$ & 01.0 & 01.0 \\
12 & $13 \pm 00$ & $\textbf{14} \pm 00$ & $\textbf{13} \pm 00$ & $\textbf{14} \pm 00$ & 01.0 & $\textbf{13} \pm 00$ & $\textbf{14} \pm 00$ & 01.0 & 01.0 \\
13 & $03 \pm 00$ & $03 \pm 00$ & $\textbf{02} \pm 00$ & $\textbf{02} \pm 00$ & 05.4 & $03 \pm 00$ & $03 \pm 00$ & 02.0 & 02.0 \\
14 & $21 \pm 00$ & $21 \pm 00$ & $\textbf{18} \pm 00$ & $\textbf{17} \pm 00$ & 11.6 & $21 \pm 00$ & $21 \pm 00$ & 02.0 & 02.0 \\
15 & $00 \pm 00$ & $00 \pm 00$ & $\textbf{00} \pm 00$ & $\textbf{00} \pm 00$ & 29.4 & $00 \pm 00$ & $00 \pm 00$ & 01.0 & 01.0 \\
16 & $09 \pm 00$ & $09 \pm 00$ & $\textbf{00} \pm 00$ & $\textbf{00} \pm 00$ & 64.0 & $09 \pm 00$ & $09 \pm 00$ & 01.0 & 01.0 \\
17 & $19 \pm 00$ & $19 \pm 00$ & $\textbf{16} \pm 00$ & $\textbf{16} \pm 00$ & 06.0 & $19 \pm 00$ & $19 \pm 00$ & 01.0 & 01.0 \\
18 & $06 \pm 00$ & $06 \pm 00$ & $\textbf{00} \pm 00$ & $\textbf{00} \pm 00$ & 64.0 & $06 \pm 00$ & $06 \pm 00$ & 01.0 & 01.0 \\
19 & $08 \pm 00$ & $08 \pm 00$ & $\textbf{00} \pm 00$ & $\textbf{00} \pm 00$ & 64.0 & $08 \pm 00$ & $08 \pm 00$ & 01.0 & 01.0 \\
20 & $01 \pm 00$ & $01 \pm 00$ & $\textbf{00} \pm 00$ & $\textbf{00} \pm 00$ & 64.0 & $01 \pm 00$ & $01 \pm 00$ & 01.0 & 01.0 \\
\hline
\textbf{avg.} & 2.35 & 2.25 & 1.40 & 1.35 & - & 2.25 & 2.40 & - & - \\
\hline
    \end{tabular}
    \caption{Comparison of QDA in terms of error in training and test sets trained with MAP, RC, and GD.}
    \label{table:results_qda_map}
\end{table*}

\begin{table*}[t]
    \centering
    \begin{tabular}{l|ccc|ccc}
    \hline
    & \multicolumn{3}{c|}{\textbf{RC vs MAP}} & \multicolumn{3}{c}{\textbf{RC vs GD}} \\
    & Better & Equal & Worse & Better & Equal & Worse \\
    \hline
    Training & 14 & 4 & 2 & 13 & 6 & 1 \\
    Test & 13 & 6 & 1 & 14 & 6 & 0 \\
    \hline
    \end{tabular}
    \caption{Summary of the comparisons of RC against MAP and GD in terms of the average training and test errors across all datasets.}
    \label{table:comparison_qda_map}
\end{table*}
In this section, we present supplementary empirical results obtained using MAP closed-form algorithms, GD, and RC with MAP for learning NB and QDA classifiers. The training and test errors obtained for NB and QDA are reported in Table~\ref{table:results_nb_map} and Table~\ref{table:results_qda_map}, respectively. A summary of the comparisons between the training and test errors for NB and QDA is provided in Table~\ref{table:comparison_nb_map} and Table~\ref{table:comparison_qda_map}, respectively. Overall, the conclusions remain consistent with those obtained when using ML as the base algorithm, as shown in Section~\ref{sec:experiments}. RC outperforms both MAP and GD in learning NB and QDA classifiers, achieving lower training and test errors.

\section{Comparison with DFE}\label{app:DFE}
\begin{table*}[t!]
    \centering
    \begin{tabular}{l|ccc|ccc|c}
    \hline
    \textbf{Index} & \multicolumn{3}{c|}{\textbf{RC}} & \multicolumn{3}{c|}{\textbf{DFE}} \\
    & Training (\%) & Test (\%)& Iter & Training (\%)& Test (\%)& Iter & Reach \\
    \hline
    1 & $\textbf{03} \pm 01$ & $\textbf{04} \pm 03$ & 41.0 & $04 \pm 01$ & $04 \pm 03$ & 03.0 & 08.6 \\
2 & $00 \pm 01$ & $03 \pm 04$ & 38.2 & $\textbf{00} \pm 00$ & $\textbf{00} \pm 00$ & 64.0 & - \\
3 & $\textbf{02} \pm 00$ & $\textbf{03} \pm 02$ & 40.2 & $03 \pm 00$ & $08 \pm 01$ & 01.0 & 01.0 \\
4 & $\textbf{30} \pm 01$ & $\textbf{36} \pm 03$ & 01.0 & $\textbf{30} \pm 01$ & $\textbf{36} \pm 03$ & 01.0 & 01.0 \\
5 & $35 \pm 01$ & $38 \pm 03$ & 01.0 & $\textbf{35} \pm 01$ & $\textbf{37} \pm 03$ & 01.8 & - \\
6 & $\textbf{22} \pm 00$ & $\textbf{24} \pm 02$ & 01.2 & $22 \pm 00$ & $25 \pm 02$ & 01.0 & 01.0 \\
7 & $\textbf{02} \pm 00$ & $\textbf{02} \pm 01$ & 64.0 & $06 \pm 00$ & $07 \pm 02$ & 09.6 & 05.4 \\
8 & $17 \pm 00$ & $20 \pm 00$ & 01.0 & $\textbf{14} \pm 00$ & $\textbf{16} \pm 01$ & 17.8 & - \\
9 & $\textbf{15} \pm 00$ & $\textbf{16} \pm 01$ & 29.2 & $17 \pm 00$ & $19 \pm 02$ & 05.4 & 03.0 \\
10 & $\textbf{41} \pm 00$ & $\textbf{44} \pm 01$ & 01.0 & $\textbf{41} \pm 00$ & $\textbf{44} \pm 01$ & 01.0 & 01.0 \\
11 & $05 \pm 00$ & $06 \pm 00$ & 07.4 & $\textbf{04} \pm 01$ & $\textbf{04} \pm 01$ & 51.4 & - \\
12 & $20 \pm 00$ & $20 \pm 01$ & 02.2 & $\textbf{16} \pm 00$ & $\textbf{16} \pm 00$ & 60.8 & - \\
13 & $\textbf{02} \pm 00$ & $\textbf{02} \pm 00$ & 09.6 & $03 \pm 00$ & $03 \pm 00$ & 01.0 & 01.0 \\
14 & $\textbf{17} \pm 00$ & $\textbf{17} \pm 00$ & 64.0 & $24 \pm 00$ & $24 \pm 00$ & 01.0 & 01.0 \\
15 & $\textbf{01} \pm 00$ & $\textbf{01} \pm 00$ & 63.2 & $01 \pm 00$ & $01 \pm 00$ & 64.0 & 25.6 \\
16 & $\textbf{07} \pm 00$ & $\textbf{07} \pm 00$ & 64.0 & $08 \pm 00$ & $08 \pm 00$ & 64.0 & 31.2 \\
17 & $\textbf{15} \pm 00$ & $\textbf{15} \pm 00$ & 09.0 & $18 \pm 00$ & $18 \pm 00$ & 02.0 & 02.0 \\
18 & $\textbf{11} \pm 00$ & $\textbf{11} \pm 00$ & 57.4 & $17 \pm 00$ & $17 \pm 00$ & 64.0 & 10.0 \\
19 & $\textbf{12} \pm 00$ & $\textbf{13} \pm 00$ & 33.2 & $17 \pm 00$ & $17 \pm 00$ & 64.0 & 08.0 \\
20 & $\textbf{04} \pm 00$ & $\textbf{04} \pm 00$ & 60.6 & $08 \pm 00$ & $09 \pm 00$ & 64.0 & 10.0 \\
\hline
    \end{tabular}
    \caption{Comparison of NB in terms of error in training and test sets trained with RC and DFE.}
    \label{table:results_dfe}
\end{table*}

\begin{table*}[t]
    \centering
    \begin{tabular}{l|ccc}
    \hline
    &\multicolumn{3}{c}{\textbf{RC vs DFE}} \\
    & Better & Equal & Worse \\
    \hline
    Training & 13 & 2 & 5 \\
    Test & 13 & 2 & 5 \\
    \hline
    \end{tabular}
    \caption{Summary of the comparisons of RC against ML and DFE in terms of the average training and test errors across all datasets.}
    \label{table:comparison_dfe}
\end{table*}

This section presents an empirical comparison between RC and Discriminative Frequency Estimate (DFE), a closely related iterative calibration procedure for generative classifiers introduced in Section~\ref{sec:dfe}. The main drawbacks of DFE are: (i) it is only applicable to classifiers with discrete variables based on Bayesian networks~\cite{bielza2014discrete}, and (ii) the equivalent sample size of the statistics increases proportionally to the training soft error with each iteration. The results are reported for RC using ML, with both RC and DFE initialized using ML.

Table~\ref{table:results_dfe} presents the results obtained on the 20 datasets using RC with ML as the closed-form learning algorithm and the Discriminative Frequency Estimate (DFE) for the discrete naïve Bayes classifier (NB). Table~\ref{table:comparison_dfe} summarizes the comparison between RC and DFE in terms of training and test errors. RC achieves lower training/test errors on 13/13 datasets, while both methods yield the same errors on 2/2 datasets out of 20.




\clearpage
\bibliographystyle{abbrv}
\bibliography{biblio}

\begin{thebibliography}{10}

\bibitem{tensorflow2015}
M.~Abadi, A.~Agarwal, P.~Barham, E.~Brevdo, Z.~Chen, C.~Citro, G.~S. Corrado, A.~Davis, J.~Dean, M.~Devin, et~al.
\newblock Tensorflow: Large-scale machine learning on heterogeneous systems, 2015.

\bibitem{augenstein2019generative}
S.~Augenstein, H.~B. McMahan, D.~Ramage, S.~Ramaswamy, P.~Kairouz, M.~Chen, R.~Mathews, and B.~A. y~Arcas.
\newblock Generative models for effective {ML} on private, decentralized datasets.
\newblock In {\em International Conference on Learning Representations}, 2020.

\bibitem{bertsekas1997nonlinear}
D.~P. Bertsekas.
\newblock Nonlinear programming.
\newblock {\em Journal of the Operational Research Society}, 48(3):334--334, 1997.

\bibitem{bielza2014discrete}
C.~Bielza and P.~Larra\~naga.
\newblock Discrete {B}ayesian network classifiers: A survey.
\newblock {\em ACM Computing Surveys}, 47(1):1--43, 2014.

\bibitem{dash2011discretization}
R.~Dash, R.~L. Paramguru, and R.~Dash.
\newblock Comparative analysis of supervised and unsupervised discretization techniques.
\newblock {\em International Journal of Advances in Science and Technology}, 2(3):29--37, 2011.

\bibitem{dempster1977}
A.~P. Dempster, N.~M. Laird, and D.~B. Rubin.
\newblock Maximum likelihood from incomplete data via the {EM} algorithm.
\newblock {\em Journal of the Royal Statistical Society: Series B}, 39(1):1--22, 1977.

\bibitem{Dua:2017}
D.~Dua and C.~Graff.
\newblock {UCI} machine learning repository, 2017.

\bibitem{TM_Lauritzen}
D.~Edwards and S.~L. Lauritzen.
\newblock {The TM algorithm for maximising a conditional likelihood function}.
\newblock {\em Biometrika}, 88(4):961--972, 12 2001.

\bibitem{elkan2001costsensitive}
C.~Elkan.
\newblock The foundations of cost-sensitive learning.
\newblock In {\em International Joint Conference on Artificial Intelligence}, volume~17, pages 973--978, 2001.

\bibitem{Elson07}
J.~Elson, J.~J. Douceur, J.~Howell, and J.~Saul.
\newblock Asirra: A captcha that exploits interest-aligned manual image categorization.
\newblock In {\em ACM Conference on Computer and Communications Security (CCS)}. Association for Computing Machinery, Inc., 2007.

\bibitem{friedman97}
N.~Friedman, D.~Geiger, and M.~Goldszmidt.
\newblock {Bayesian Network Classifiers}.
\newblock {\em Machine Learning}, 29(2--3):131--163, 1997.

\bibitem{GM20}
H.~GM, M.~K. Gourisaria, M.~Pandey, and S.~S. Rautaray.
\newblock A comprehensive survey and analysis of generative models in machine learning.
\newblock {\em Computer Science Review}, 38:100285, 2020.

\bibitem{greiner02}
R.~Greiner, X.~Su, B.~Shen, and W.~Zhou.
\newblock Structural extension to logistic regression: Discriminative parameter learning of belief net classifiers.
\newblock {\em Machine Learning}, 59:297--322, 2005.

\bibitem{hastie2009}
T.~Hastie, R.~Tibshirani, and J.~Friedman.
\newblock {\em The Elements of Statistical Learning: Data Mining, Inference, and Prediction}.
\newblock Springer, 2009.

\bibitem{He16}
K.~He, X.~Zhang, S.~Ren, and J.~Sun.
\newblock Deep residual learning for image recognition.
\newblock In {\em IEEE Conference on Computer Vision and Pattern Recognition}, pages 770--778, 2016.

\bibitem{Jebara12}
T.~Jebara.
\newblock {\em Machine Learning: Discriminative and Generative}.
\newblock Springer New York, NY, 2012.

\bibitem{Kim23}
S.~Kim, H.~Kim, E.~Yun, H.~Lee, J.~Lee, and J.~Lee.
\newblock Probabilistic imputation for time-series classification with missing data.
\newblock In {\em Proceedings of the 40th International Conference on Machine Learning}, ICML'23. JMLR.org, 2023.

\bibitem{Krizhevsky09}
A.~Krizhevsky.
\newblock Learning multiple layers of features from tiny images.
\newblock Technical report, 2009.

\bibitem{lecun2010mnist}
Y.~LeCun, C.~Cortes, and C.~Burges.
\newblock {MNIST} handwritten digit database.
\newblock {\em ATT Labs}, 2, 2010.

\bibitem{Murphy12}
K.~P. Murphy.
\newblock {\em Machine Learning: A Probabilistic Perspective}.
\newblock The MIT Press, 2012.

\bibitem{ng2001discriminative}
A.~Ng and M.~Jordan.
\newblock On discriminative vs. generative classifiers: A comparison of logistic regression and naive {B}ayes.
\newblock In {\em Advances in Neural Information Processing Systems}, volume~14, 2001.

\bibitem{perez2006gaussian}
A.~P{\'e}rez, P.~Larra{\~n}aga, and I.~Inza.
\newblock Supervised classification with conditional {G}aussian networks: Increasing the structure complexity from naive {B}ayes.
\newblock {\em International Journal of Approximate Reasoning}, 43(1):1--25, 2006.

\bibitem{pernkopf2005discriminative}
F.~Pernkopf and J.~Bilmes.
\newblock Discriminative versus generative parameter and structure learning of {B}ayesian network classifiers.
\newblock In {\em International Conference on Machine Learning}, pages 657--664, 2005.

\bibitem{roos2005discriminative}
T.~Roos, H.~Wettig, P.~Gr{\"u}nwald, P.~Myllym{\"a}ki, and H.~Tirri.
\newblock On discriminative {B}ayesian network classifiers and logistic regression.
\newblock {\em Machine Learning}, 59:267--296, 2005.

\bibitem{Guzman07}
G.~Santaf{\'e}, J.~A. Lozano, and P.~Larra{\~{n}}aga.
\newblock Discriminative vs. {G}enerative {L}earning of {B}ayesian {N}etwork {C}lassifiers.
\newblock In {\em Symbolic and Quantitative Approaches to Reasoning with Uncertainty}, pages 453--464. Springer Berlin Heidelberg, 2007.

\bibitem{Tolou23}
T.~Shadbahr, M.~Roberts, J.~Stanczuk, J.~Gilbey, P.~Teare, S.~Dittmer, M.~Thorpe, R.~Torn{\'e}, E.~Sala, P.~Li{\'o}, M.~Patel, J.~Preller, I.~Selby, A.~Breger, J.~Weir-McCall, E.~Gkrania-Klotsas, A.~Korhonen, E.~Jefferson, G.~Langs, G.~Yang, H.~Prosch, J.~Babar, L.~{Escudero S{\'a}nchez}, M.~Wassin, M.~Holzer, N.~Walton, J.~Rudd, T.~Mirtti, A.~Rannikko, J.~Aston, J.~Tang, and C.-B. Sch{\"o}nlieb.
\newblock The impact of imputation quality on machine learning classifiers for datasets with missing values.
\newblock {\em Communications Medicine}, 3(1), Oct. 2023.

\bibitem{su2008discriminative}
J.~Su, H.~Zhang, C.~X. Ling, and S.~Matwin.
\newblock Discriminative parameter learning for {B}ayesian networks.
\newblock In {\em International Conference on Machine Learning}, pages 1016--1023, 2008.

\bibitem{Cuesta2019}
Y.~Sun, A.~Cuesta-Infante, and K.~Veeramachaneni.
\newblock Learning vine copula models for synthetic data generation.
\newblock {\em AAAI Conference on Artificial Intelligence}, 33(01):5049--5057, July 2019.

\bibitem{sundberg2002convergence}
R.~Sundberg.
\newblock The convergence rate of the tm algorithm of edwards \& lauritzen.
\newblock {\em Biometrika}, 89(2):478--483, 2002.

\bibitem{vapnik2013nature}
V.~Vapnik.
\newblock {\em The nature of statistical learning theory}.
\newblock Springer {S}cience \& {B}usiness media, 2013.

\bibitem{Xiao17}
H.~Xiao, K.~Rasul, and R.~Vollgraf.
\newblock Fashion-{MNIST}: a novel image dataset for benchmarking machine learning algorithms.
\newblock {\em CoRR}, abs/1708.07747, 2017.

\bibitem{Zheng23}
C.~Zheng, G.~Wu, F.~Bao, Y.~Cao, C.~Li, and J.~Zhu.
\newblock Revisiting discriminative vs. generative classifiers: theory and implications.
\newblock In {\em International Conference on Machine Learning}, 2023.

\end{thebibliography}

\end{document}